\documentclass{article}
\usepackage{spconf}
\usepackage{amsmath}
\usepackage{amssymb}
\usepackage{booktabs}
\usepackage{microtype}
\usepackage{graphicx}
\usepackage{cite}
\usepackage{dcolumn}
\usepackage{nth}
\usepackage{paralist}
\usepackage[bookmarks=false]{hyperref}
\hypersetup{pdftitle={DESIGN OF THE IBM GRANITE 5.0 TURBOCTC ASR MODEL},pdfauthor={Brian Kingsbury, George Saon, Masayuki Suzuki, Jeff Kuo, Takashi Fukuda, Samuel Thomas, Vishal Sunder, and Avihu Dekel}}
\newcolumntype{d}[1]{D{.}{.}{#1}}
\newcommand{\turboctc}{\texttt{granite\--speech\--5.0\--470m\--turboctc}}
\newcommand{\turboctcnc}{\texttt{granite\--speech\--5.0\--470m\--turboctc\--nc}}
\title{DESIGN OF THE IBM GRANITE 5.0 TURBOCTC ASR MODEL}
\name{\shortstack{Brian Kingsbury, George Saon, Masayuki Suzuki, Hong-Kwang J. Kuo\\Takashi Fukuda, Samuel Thomas, Vishal Sunder, Avihu Dekel}}
\address{IBM Research}
\begin{document}
\ninept
\maketitle
\begin{abstract}
We describe the architecture, training methodology and inference speedups of Granite 5.0 Turbo CTC, a 470 million parameter encoder-only model with an excellent speed-accuracy tradeoff. The architecture uses pyramidal temporal subsampling within Conformer blocks using strided depthwise convolutions, block-diagonal (chunk-wise) self-attention, and conditioning on intermediate predictions from the middle layer. Training highlights are the use of only publicly available data, the novel use of a Muon optimizer, and balanced data sampling. Inference speedups include replacing $1 \times 1$ convolutions with linear layers and optimizing the attention computation in the Conformer blocks. Collectively, these result in a model that is on the speed-accuracy Pareto frontier of the Open ASR leaderboard for English short-form ASR while being twice as fast as the fastest competitor. The model can be used under a permissive license and downloaded from \url{https://huggingface.co/ibm-granite/granite-speech-5.0-470m-turboctc}. 
\end{abstract}

\section{Introduction}
\label{sec:intro}
Recent improvements in automatic speech recognition (ASR) have been driven by the integration of text-based large language models (LLMs) into ASR systems. One popular approach to integration uses a separately trained acoustic encoder, connects the encoder to the LLM via a trainable projector, and equips the LLM with low rank adapters~\cite{Tang2024,Ma2025}. Such speech language models enjoy capabilities like speech translation, contextual biasing, and seamless integration of dialog context as well as strong transcription accuracy~\cite{grattafiori2024llama,chen2024salm,abouelenin2025phi,Saon2025,shi2026qwen3}.

\begin{figure}
    \centerline{\includegraphics[width=0.90\columnwidth]{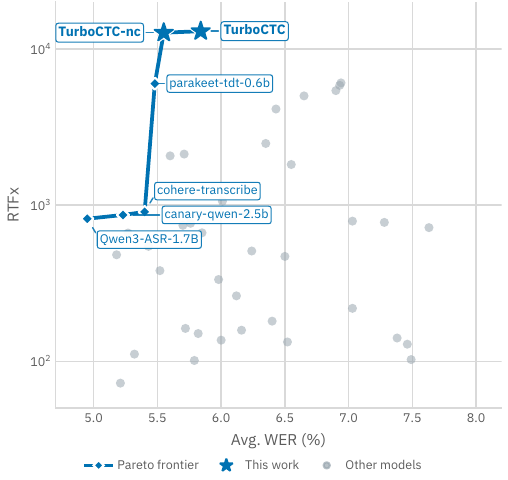}}
    \caption{\label{fig:pareto} OpenASR leaderboard speed-accuracy Pareto curve as of 4 September 2026. Higher RTFx and lower WER are better; \turboctc{} and \turboctcnc{} are marked with stars. The axes are cropped for legibility, placing 6 of the 51 models outside the view; the entire Pareto frontier is shown.}
\end{figure}

However, LLM integration is costly because the LLM brings in a large number of parameters, driving up memory footprint, and the autoregressive generation process used in most LLMs limits transcription speed. Moreover, a well-trained Conformer model is a strong ASR model in its own right~\cite{ng2021pushing}. Therefore, in this paper we focus on encoder-only speech recognition using a Conformer~\cite{Gulati2020} model trained using the connectionist temporal classification (CTC)~\cite{Graves2006} loss. We make the following contributions in this paper.
\begin{itemize}
    \item We use strided depthwise convolutions in the Conformer blocks to implement pyramidal temporal subsampling.
    \item We apply Muon with Polar Express orthogonalization to train the base model.
    \item We accelerate inference by replacing $1 \times 1$ convolutions with linear layers and optimizing the attention computations in the Conformer blocks.
    \item We improve the accuracy of the model by distilling knowledge from an LLM and by performing robust fine tuning.
\end{itemize}
These contributions are verified in a strong model that was the fastest on the OpenASR Leaderboard~\cite{OpenASR} as of 4 September 2026 and resided on the speed-accuracy Pareto frontier as shown in Figure~\ref{fig:pareto}. A companion model that had additional training data, \turboctcnc{}, was only 0.7\% behind the top model, a speech language model, in accuracy. Owing to their encoder-only design, both models are extremely small, requiring only 470 million parameters.

\section{Architecture}
\label{sec:architecture}
In this section we detail the architectural design elements for \turboctc{} pertaining to temporal downsampling, block attention and conditioning on intermediate predictions. The model consists of 16 Conformer blocks with a hidden dimension of 1024 and an output layer of size 16,384 for a total of 470 million parameters. Other architectural details are shown in Table~\ref{tab:arch}.

\begin{table}[htb]
\begin{center}
\begin{tabular}{lc} \toprule
{\bf Configuration parameter} & {\bf Value}\\ \midrule
Input dimension & 320 (80 logmels + 80 deltas) x 2\\ 
Number of layers   & 16                   \\ 
Hidden dimension & 1024                \\ 
Number of attention heads & 8             \\ 
Attention head size    & 128           \\ 
Attention block size   & 128           \\ 
Convolution kernel size & 7           \\ 
Output dimension        & 16384           \\ \bottomrule
\end{tabular}
\end{center}
\caption{\label{tab:arch} Configuration parameters for the model architecture.}
\end{table}

\begin{figure}
    \centerline{\includegraphics[width=\columnwidth]{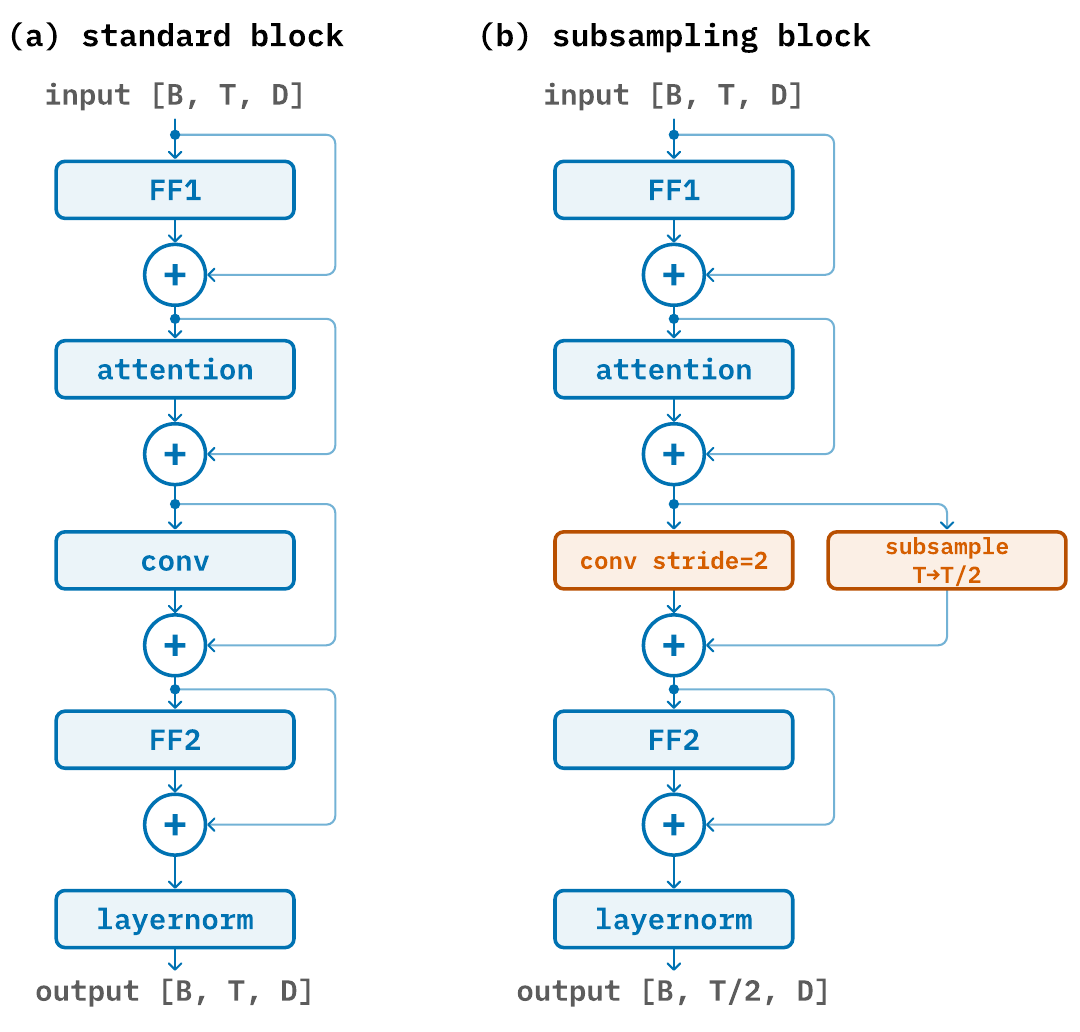}}
    \caption{\label{fig:subsampling} Comparison of standard and subsampling Conformer blocks.}
\end{figure}

Temporal downsampling is achieved in two separate steps. In step 1, we reduce the log-Mel frame rate from 100 Hz to 50 Hz by stacking every two consecutive frames and skipping every other frame through a simple tensor reshape operation. In step 2, the first two Conformer blocks perform temporal downsampling, each achieving a 2x temporal reduction (total of 4x) resulting in a 12.5 Hz frame/token rate that gets propagated through the remaining Conformer blocks and output layer. Inspired by~\cite{burchi2021efficient}, we change the depthwise convolutions in the Conformer blocks to stride 2 convolutions. In order to match the sequence length of the residual coming out of the self-attention layer to the resulting half-length sequence from the convolutional layer, we also perform average pooling of every 2 consecutive residual attention frames. The difference between standard and subsampling Conformer blocks is illustrated in Figure~\ref{fig:subsampling}. To compensate for the reduced token rate, we use an output layer with 16K BPE subword units derived from the acoustic transcripts. In Table~\ref{tab:subsample}, we compare our subsampling approach with purely convolutional subsampling as proposed in the FastConformer architecture~\cite{rekesh2023fast} on the public English test sets of the Open ASR leaderboard for the same 8x temporal reduction (rows 1 and 2).

Secondly, we opt for block-diagonal (or chunkwise) self-attention in the Conformer layers. This has two advantages over full attention: (i) the complexity scales linearly instead of quadratically with the length of the audio sequence and (ii) it generalizes better to long-form audio based on the findings of~\cite{GoogleUSM}. In practice, we use chunk sizes of 128 frames for all Conformer layers (including the first two subsampling layers) which corresponds to 2.5 seconds (layer 1), 5 seconds (layer 2) and 10 seconds (layers 2-16) of audio. In the same Table~\ref{tab:subsample}, we compare block attention with grouped attention proposed in~\cite{burchi2021efficient} where for the latter we used 3 groups for the first 6 conformer layers (rows 1 and 3).


Lastly, we use conditioning on intermediate predictions from the middle Conformer layer as proposed in~\cite{nozaki2021relaxing}. The frame-level softmax distributions from the 8th layer are linearly mapped back to the hidden dimension and added as a residual connection to the input of the 9th layer. The output layers are tied between the middle and last layers and the CTC loss is computed as a weighted sum between the intermediate and last layer CTC losses with weights 0.2 and 0.8, respectively. The importance of self-conditioning is shown in Table~\ref{tab:subsample} (rows 1 and 4) where, if we train a model without it, the performance degrades across all test sets.

\begin{table*}
  \centering
  \begin{tabular}{ld{1.2}d{1.2}d{2.2}d{1.2}d{1.2}d{1.2}d{1.2}d{1.2}}\toprule
    {\bf Architecture} & \multicolumn{1}{c}{\bf AMI} & \multicolumn{1}{c}{\bf Earnings22} & \multicolumn{1}{c}{\bf Gigaspeech} & \multicolumn{1}{c}{\bf LS-clean} & \multicolumn{1}{c}{\bf LS-other} & \multicolumn{1}{c}{\bf SPGI} & \multicolumn{1}{c}{\bf VoxPopuli} & \multicolumn{1}{c}{\bf AVG}\\\midrule
    Proposed            & 8.53 & 8.65 & 10.17 & 1.34 & 2.63 & 3.37 & 6.13 & 5.83 \\
    Convolutional front end~\cite{rekesh2023fast} & 8.71 & 8.68 & 10.22 & 1.37 & 2.72 & 3.58 & 6.15 & 5.92 \\
    Grouped attention~\cite{burchi2021efficient}  & 8.96 & 8.79 & 10.21 & 1.37 & 2.69 & 3.38 & 6.15 & 5.94 \\
    No self-conditioning~\cite{nozaki2021relaxing} & 8.89 & 9.72 & 10.28 & 1.46 & 2.85 & 3.55 & 6.25 & 6.00 \\\bottomrule
  \end{tabular}
  \caption{\label{tab:subsample} WER performance on the public components of the OpenASR Leaderboard for different approaches to temporal subsampling, attention and self-conditioning.}
\end{table*}
\section{Training}
\label{sec:training}
\begin{table*}
  \centering
  \begin{tabular}{ld{1.2}d{1.2}d{2.2}d{1.2}d{1.2}d{1.2}d{1.2}d{1.2}}\toprule
    {\bf Training stage} & \multicolumn{1}{c}{\bf AMI} & \multicolumn{1}{c}{\bf Earnings22} & \multicolumn{1}{c}{\bf Gigaspeech} & \multicolumn{1}{c}{\bf LS-clean} & \multicolumn{1}{c}{\bf LS-other} & \multicolumn{1}{c}{\bf SPGI} & \multicolumn{1}{c}{\bf VoxPopuli} & \multicolumn{1}{c}{\bf AVG}\\\midrule
    Base                & 8.53 & 8.65 & 10.17 & 1.34 & 2.63 & 3.37 & 6.13 & 5.83 \\
    LLM distillation    & 8.51 & 8.62 &  9.95 & 1.33 & 2.56 & 3.28 & 5.98 & 5.75 \\
    Robust fine tuning  & 8.32 & 8.54 &  9.93 & 1.32 & 2.54 & 3.35 & 6.03 & 5.72 \\\bottomrule
  \end{tabular}
  \caption{\label{tab:training} WER performance on the public components of the OpenASR Leaderboard for the three training stages.}
\end{table*}
The model is trained in three phases:
\begin{inparaenum}[(1)]
  \item base model training with the CTC loss,
  \item knowledge distillation from a Granite LLM, and
  \item robust fine tuning.
\end{inparaenum}
Table~\ref{tab:training} shows how each stage of training improves the accuracy of the model. The following subsections describe these stages in more detail.
\subsection{Base model training}
\label{sec:muon}
\begin{table}
  \centering
  \begin{tabular}{llr}\toprule
    {\bf Dataset} & {\bf Source} & \multicolumn{1}{c}{\bf \# hours}\\\midrule
    MLS~\cite{Pratap2020} & natural & 44600\\
    YODAS~\cite{Li2023} & natural & 8900\\
    CommonVoice-17~\cite{Ardila2020} & natural & 2500\\
    Librispeech~\cite{Panayotov2015} & natural & 960\\
    VoxPopuli~\cite{Wang2021}  & natural & 500\\
    AMI~\cite{McCowan2005} & natural & 150\\
    Earnings-22~\cite{DelRio2022} & natural & 100\\
    Multispeaker & synthetic & 2000\\
    Multispeaker-Earnings & synthetic & 500\\
    Numbers & synthetic & 240\\\bottomrule
  \end{tabular}
  \caption{\label{tab:data} List of datasets used to train \turboctc{}.}
\end{table}
The model is trained on the approximately 60k hours of audio from publicly available English ASR corpora and synthetic data listed in Table~\ref{tab:data}. {\bf Multispeaker} is synthesized by concatenating single-speaker segments sampled from MLS, YODAS, CommonVoice-17, VoxPopuli, and AMI. {\bf Multispeaker-Earnings} is synthesized by concatenating single-speaker segments sampled from Earnings-22. {\bf Numbers} comprises a collection of utterances containing numeric expresssions, phone numbers, monetary values, e-mail addresses, URLs, and street addresses. The utterances were generated using \texttt{gpt-oss-120b}~\cite{GPTOSS} or \texttt{gpt-oss-20b} and synthesized with a StyleTTS 2~\cite{li2023a} model.

Initial model development was done using AdamW~\cite{Loshchilov2019}, but as described below we switched to a Muon~\cite{Jordan2024} variant for training the final model. Other training hyperparameters were set as follows.
\begin{itemize}
  \item Training on a compute node with 8 NVIDIA H100 GPUs.
  \item Balanced data sampling~\cite{Saon2025} with balancing parameter $\alpha = 0.8$ and up to 448 seconds of audio per GPU.
  \item Weight initialization uses PyTorch 2.11 defaults except for the output biases and intermediate projection weights and biases, which are initialized to $0.0$.
  \item PyTorch \texttt{OneCycleLR} learning rate schedule, linear warmup from $5 \times 10^{-5}$ to $5 \times 10^{-4}$ over 177k updates, linear annealing to $1.5 \times 10^{-7}$ over 1.71M updates, and default momentum cycling on both Muon's momentum and AdamW's $\beta$.
  \item The gradient $L2$ norm is clipped to $10.0$.
  \item Weight decay with $\lambda= 0.01$ is applied in AdamW and Muon, excluding biases and normalization parameters.
  \item Dropout~\cite{Srivastava2014} is applied inside the Conformer blocks with $p = 0.1$ to the outputs of the multihead attention, convolution, and feedforward blocks, with $p = 0.1$ to the hidden layer in the feedforward blocks, and with $p = 0.25$ before the output projection.
  \item SpecAugment~\cite{Park2019} is applied to the input features. The frequency masks are up to 15 filters, the number of frequency masks is sampled uniformly from $\{1, 2\}$, and with $p = 0.10$ no frequency masking is applied. The time masks are up to 24 frames long and the number of time masks is sampled uniformly from $\{0, \dotsc, \lfloor 0.03125 * T \rfloor\}$, where $T$ is the length of the utterance in frames. Identical time and frequency masking are applied to the static and delta feature streams.
  \item Noise augmentation is applied to the input audio. We add noise samples with $p = 0.25$ where the signal-to-noise ratio is sampled from $\mathcal{U}[-5, 20]$ dB.
\end{itemize}

We investigated three different optimization algorithms for training the base model:
\begin{inparaenum}[(1)]
  \item AdamW,
  \item Muon, and
  \item Muon with Polar Express (MuonP for short)~\cite{Amsel2026}.
\end{inparaenum}
Muon, which has recently been used to train frontier LLMs~\cite{Liu2025}, optimizes {\bf matrix-structured} parameters via steepest descent in the spectral norm~\cite{Bernstein2024}, relying on iterative methods to approximate the necessary polar decomposition of the update matrix. Because Muon and MuonP apply only to two-dimensional parameters that act as linear operators, we rely on AdamW to train the biases, normalization parameters, relative positional embeddings, temporal convolutions, output projection, and the map from intermediate token posteriors back to the internal dimension of the Conformer stack. We use the update scaling and weight decay from \cite{Liu2025}.

\begin{table*}
  \centering
  \begin{tabular}{ld{1.2}d{1.2}d{2.2}d{1.2}d{1.2}d{1.2}d{1.2}d{1.2}}\toprule
    {\bf optimizer} & \multicolumn{1}{c}{\bf AMI} & \multicolumn{1}{c}{\bf Earnings22} & \multicolumn{1}{c}{\bf Gigaspeech} & \multicolumn{1}{c}{\bf LS-clean} & \multicolumn{1}{c}{\bf LS-other} & \multicolumn{1}{c}{\bf SPGI} & \multicolumn{1}{c}{\bf VoxPopuli} & \multicolumn{1}{c}{\bf AVG}\\\midrule
    AdamW & 8.66 & 8.89 & 10.21 & 1.37 & 2.75 & 3.65 & 6.41 & 5.99 \\
    Muon  & 8.67 & 8.88 & 10.04 & 1.40 & 2.65 & 3.43 & 6.44 & 5.93 \\
    MuonP & 8.53 & 8.65 & 10.17 & 1.34 & 2.63 & 3.37 & 6.13 & 5.83 \\\bottomrule
  \end{tabular}
  \caption{\label{tab:muon} WER performance on the public components of the OpenASR Leaderboard for different optimizers.}
\end{table*}
Table~\ref{tab:muon} compares the performance of the three optimizers on the public parts of the 20 May 2026 version of the OpenASR leaderboard. All hyperparameter setttings are as listed above for all three training runs. Both Muon variants outperform AdamW in aggregate word error rate, and the Polar Express version of Muon outperforms AdamW on all test sets. The MuonP-trained Conformer thus became the base for \turboctc{}.

\subsection{LLM Knowledge Distillation}
\label{sec:distillation}
We train an auxiliary 4-layer, 1024 hiddens transformer decoder with 75 million parameters used to predict the output distribution of the {\tt granite-4.1-8b-base} text LLM. The decoder cross-attends to the encoder embeddings from the last layer and is conditioned on the previous token embeddings with teacher forcing during training. The distillation loss measures the KL divergence between the per-token teacher softmax distribution truncated to the top 10 tokens and the student distribution where the latter shares the frozen {\tt granite-4.1-8b-base} output layer of size 100 K BPE units. Importantly, the AED decoder is only used for training and discarded at inference. Training is done with the MuonP optimizer on the acoustic training data from Table~\ref{tab:data} over 5 epochs with linear annealing using separate maximum learning rates for the CTC pre-trained encoder (1e-6) and AED decoder (1e-3). The combined training loss is a weighted sum of the per-frame CTC loss and the distillation loss with weights of 0.3 and 0.7, respectively. The improvements in accuracy due to this step are shown in Table~\ref{tab:training} (rows 1 and 2).

\subsection{Robust Fine-Tuning}
\label{sec:robustness}

To improve robustness to acoustic distortions while preserving recognition accuracy on clean speech, we combine multi-condition fine-tuning on paired clean and acoustically perturbed speech with parameter interpolation between the original and adapted models. For each training utterance $x$ with transcription $y$, we first apply speed perturbation, sampling the factor uniformly from $\{0.8,0.9,1.0,1.1,1.2\}$. The resulting waveform serves as the clean view $x_c$, and a copy is further perturbed to produce the distorted view $x_d$. Both views share the same transcription. The clean view receives no added noise, reverberation, bandwidth filtering, or codec processing.

To construct the distorted view, we independently apply additive background noise, reverberation, bandwidth filtering, and codec and quantization processing with probabilities $0.25$, $0.50$, $0.30$, and $0.15$, respectively. When noise is applied, the SNR is sampled uniformly from $[-5,20]$~dB. Noise recordings are drawn from the noise subset of MUSAN~\cite{snyder2015musan}, which contains recordings originating from Freesound and SoundBible. Each noise recording is randomly cropped, with repetition where necessary, to match the utterance duration.

Reverberation is generated by convolving the waveform with measured or simulated room impulse responses from the Room Impulse Response and Noise Database (OpenSLR SLR28)~\cite{ko2017reverberation}, including responses from the REVERB Challenge 2014 dataset and the Aachen Impulse Response Database. These responses span a range of acoustic conditions, including far-microphone configurations, large rooms, lecture rooms, meeting rooms, offices, and a highly reverberant hall. The convolved waveform is cropped to the original duration and rescaled to preserve its pre-convolution peak amplitude.

Bandwidth perturbations use high-pass, low-pass, and band-pass filters, including telephone-band filtering at 300--3400~Hz and more restrictive conditions. Channel perturbations include PCM quantization, $\mu$-law, A-law, GSM, and ADPCM processing, together with low-pass filtering and quantization to approximate additional lossy channel effects.

The model is trained on the paired views using a weighted combination of CTC losses:
\begin{equation}
  \mathcal{L}_{\mathrm{robust}}(\theta;x,y)
  = (1-w)\mathcal{L}_{\mathrm{CTC}}(\theta;x_c,y)
  + w\mathcal{L}_{\mathrm{CTC}}(\theta;x_d,y),
\end{equation}
where $w \in [0,1]$ controls the relative contribution of the distorted-view loss. We select $w=0.5$ for the final configuration. The clean-view loss helps preserve recognition accuracy on clean speech, while the distorted-view loss encourages robustness to acoustic distortions. We also experimented with an additional consistency loss between the predictions for the clean and distorted views, but omitted it from the final objective because it did not improve recognition performance.

We fine-tune the full CTC model for one epoch on the same speech training data using the combined loss from the clean and distorted views. The initial learning rate is $10^{-5}$ and is linearly annealed to $10^{-7}$ after a short initial plateau. Balanced sampling uses a balancing parameter of $0.9$; the optimizer configuration and other regularization settings are retained from base model training.

Finally, to balance robustness gains with the preservation of clean-speech recognition accuracy, we interpolate the model parameters before and after robust fine-tuning:
\begin{equation}
  \theta_{\mathrm{interp}}
  = (1-\gamma)\theta_0 + \gamma\theta_r,
\end{equation}
where $\theta_0$ denotes the model before adaptation and $\theta_r$ the robust fine-tuned model, with $\gamma=0.4$ in practice. Interpolation produces a single model with unchanged architecture and inference cost.

Robust fine-tuning contributed to our models being ranked in the top 10 by accuracy on the noisy and far-field ASR (FFASR) leaderboard~\cite{FFASR} while having the two highest inference speeds.
\section{Speedups}
\label{sec:speedups}
\begin{table}
  \centering
  \begin{tabular}{ccd{5.0}d{2.0}}\toprule
    {\bf efficient?} & {\bf linear?} & \multicolumn{1}{c}{\bf RTFx} & \multicolumn{1}{c}{\bf speedup (\%)} \\\midrule
    $\times$     & $\times$     &  9229 &  0 \\
    $\checkmark$ & $\times$     & 11369 & 23 \\
    $\times$     & $\checkmark$ & 10688 & 16 \\
    $\checkmark$ & $\checkmark$ & 13715 & 49 \\\bottomrule
  \end{tabular}
  \caption{\label{tab:speedups} Inverse real time factors for the baseline model and models using one or both of our speedups.}
\end{table}
Our Conformer code is based on a well-known open-source implementation~\cite{lucidrains} with several modifications. We implemented chunked attention~\cite{GoogleUSM} to avoid quadratic memory and compute scaling, we replaced the \texttt{einsum}-based content attention with PyTorch's \texttt{scaled\_dot\_product\_attention}, and we implemented masking to prevent padded locations from affecting the model's output.

On top of this baseline, we identified two code changes that accelerated training and inference. First, we observed that, due to the way the attention operation was organized, the \texttt{MATH} backend was always invoked. Reorganization of this computation permitted consistent invocation of the \texttt{EFFICIENT\_ATTENTION} backend instead. Note that the use of Shaw's relative positional encoding precludes the use of FlashAttention. Second, we replaced the $1 \times 1$ convolutions in the Conformer's depthwise convolution block with mathematically equivalent \texttt{Linear} operations after we found that the corresponding kernels were more efficient.

Table~\ref{tab:speedups} reports inverse real time factors (RTFx) for the public datasets on the 20 May 2026 version of the OpenASR leaderboard. The measurements were made with a batch size of 128 utterances using a single NVIDIA H100 GPU on a Dell XE9680 node running PyTorch 2.6.0 with CUDA 12.4. These tests were done using a model checkpoint from the end of the first phase of training, prior to LLM knowledge distillation and robust fine tuning, with greedy decoding. Each of the four model runs achieves an aggregate 5.83\% WER. Taken together, the efficient attention kernel and use of \texttt{Linear} operations instead of \texttt{Conv1d} operations speed up inference by 49\%, with the larger gain coming from the efficient attention. We observed less dramatic speedups in training because that process involves other operations that are not accelerated by these changes.

\section{Conclusions}
\label{sec:conclusions}
We have described \turboctc{} and shown how specific architectural choices, self-conditioning and temporal dowsampling via strided depthwise convolutions in the first two conformer blocks; training methods, the use of Muon for base training, distillation from a Granite LLM, and robust fine tuning; and inference speedups, replacing $1 \times 1$ convolutions with linear layers and optimizing attention kernels, combine to produce an ASR model that is on the speed-accuracy Pareto frontier of the Open ASR leaderboard for English short-form ASR while being twice as fast as the fastest competitor.

\section{Use of AI}
Claude Code was used to aid in writing portions of the training and inference code and in the preparation of Figures~\ref{fig:pareto} and~\ref{fig:subsampling}. All experiments in this paper were run and verified by hand. The authors are fully responsible for the content of this submission.

\bibliographystyle{IEEEbib_short}
\bibliography{refs}

@InProceedings{Amsel2026,
  author =       {N. Amsel and D. Persson and C. Musco and
                  R. M. Gower},
  title =        {The {Polar} {Express}: Optimal matrix sign methods
                  and their application to the {Muon} algorithm},
  booktitle =    {Proc. ICLR},
  year =         2026
}

@InProceedings{Ardila2020,
  author =       {R. Ardila and M. Branson and K. Davis and
                  M. Henretty and M. Kohler and J. Meyer and R. Morais
                  and L. Saunders and F. M. Tyers and G. Weber},
  title =        {{Common} {Voice}: A Massively-Multilingual Speech
                  Corpus},
  booktitle =    {Proc. LREC},
  year =         2020
}

@InProceedings{Bernstein2024,
  author =       {J. Bernstein and L. Newhouse},
  title =        {Old optimizer, new norm: an anthology},
  booktitle =    {OPT2024: 16th Annual Workshop on Optimization for
                  Machine Learning},
  year =         2024
}

@misc{DelRio2022,
  title =        {Earnings-22: A Practical Benchmark for Accents in
                  the Wild},
  author =       {Del Rio, M. and P. Ha and Q. McNamara and C. Miller
                  and S. Chandra},
  howpublished = {\url{https://arxiv.org/pdf/2203.15591}},
  year =         2022
}

@misc{GPTOSS,
  title =        {{gpt-oss-120b} and {gpt-oss-20b} model card},
  author =       {OpenAI},
  howpublished = {\url{https://arxiv.org/abs/2508.10925}},
  year =         2025
}

@misc{GoogleUSM,
  title =        {Google {USM}: Scaling automatic speech recognition
                  beyond 100 languages},
  author =       {Y. Zhang and W. Han and J. Qin and Y. Wang and
                  A. Bapna and Z. Chen and N. Chen and B. Li and
                  V. Axelrod and G. Wang and Z. Meng and K. Hu and
                  A. Rosenberg and R. Prabhavalkar and D. S. Park and
                  P. Haghani and J. Riesa and G. Perng and H. Soltau
                  and T. Strohman and B. Ramabhadran and T. Sainath
                  and P. Moreno and C.-C. Chiu and J. Schalkwyk and
                  F. Beaufays and Y. Wu},
  howpublished = {\url{https://arxiv.org/abs/2303.01037}},
  year =         2023,
}

@InProceedings{Graves2006,
  author =       {A. Graves and S. Fern\'{a}ndez and F. Gomez and
                  J. Schmidhuber},
  title =        {Connectionist temporal classification: labelling
                  unsegmented sequence data with recurrent neural
                  networks},
  booktitle =    {Proc. ICML},
  year =         2006
}

@InProceedings{Gulati2020,
  author =       {A. Gulati and J. Qin and C. Chiu and N. Parmar and
                  Y. Zhang and J. Yu and W. Han and S. Wang and
                  Z. Zhang and Y. Wu and R. Pang},
  title =        {Conformer: Convolution-augmented Transformer for
                  Speech Recognition},
  booktitle =    {Proc. INTERSPEECH},
  year =         2020
}

@Misc{Jordan2024,
  author =       {K. Jordan},
  title =        {Muon: an optimizer for hidden layers in neural
                  networks},
  howpublished = {\url{https://kellerjordan.github.io/posts/muon/}}
}

@InProceedings{Li2023,
  author =       {X.Li and S. Takamichi and T. Saeki and W. Chen and
                  S. Shiota and S. Watanabe},
  booktitle =    {Proc. ASRU},
  title =        {{YODAS}: {YouTube}-Oriented Dataset for Audio and
                  Speech},
  year =         2023
}

@InProceedings{Li2023a,
  author =       {Y. A. Li and C. Han and V. S. Raghavan and
                  G. Mischler and N. Mesgarani},
  title =        {{StyleTTS} 2: Towards Human-Level Text-to-Speech
                  through Style Diffusion and Adversarial Training
                  with Large Speech Language Models},
  booktitle =    {Proc. NeurIPS},
  year =         2023
}

@Misc{Liu2025,
  author =       {J. Liu and J. Su and X. Yao and Z. Jiang and G. Lai
                  and Y. Du and Y. Qin and W. Xu and E. Lu and J. Yan
                  and Y. Chen and H. Zheng and Y. Liu and S. Liu and
                  B. Yin and W. He and H. Zhu and Y. Wang and J. Wang
                  and M. Dong and Z. Zhang and Y. Kang and H. Zhang
                  and X. Xu and Y. Zhang and Y. Wu and X. Zhou and
                  Z. Yang},
  title =        {Muon is scalable for {LLM} training},
  howpublished = {\url{https://arxiv.org/pdf/2502.16982}},
  year =         2025
}

@InProceedings{Loshchilov2019,
  author =       {I. Loshchilov and F. Hutter},
  title =        {Decoupled weight decay regularization},
  booktitle =    {Proc. ICLR},
  year =         2019
}

@InProceedings{Ma2025,
  author =       {Z. Ma and G. Yang and Y. Yang and Z. Gao and J. Wang
                  and Z. Du and F. Yu and Q. Chen and S. Zheng and
                  S. Zhang and X. Chen},
  title =        {Speech recognition meets large language model:
                  Benchmarking, models, and exploration},
  booktitle =    {Proc. AAAI},
  year =         2025
}

@InProceedings{McCowan2005,
  title =        "The {AMI} meeting corpus",
  author =       "I. McCowan and J. Carletta and W. Kraaij and
                  S. Ashby and S. Bourban and M. Flynn and
                  M. Guillemot and T. Hain and J. Kadlec and
                  V. Karaiskos and M. Kronenthal and G. Lathoud and
                  M. Lincoln and A. Lisowska and W. Post and
                  D. Reidsma and P. Wellner",
  year =         2005,
  booktitle =    "Proc. Measuring Behavior"
}

@misc{OpenASR,
    title = {Open {ASR} {Leaderboard}},
    howpublished={\url{https://huggingface.co/spaces/hf-audio/open_asr_leaderboard}},
    note = {Accessed 2 September 2026}
}

@misc{FFASR,
    title = {Far-{F}ield {ASR} {Leaderboard}},
    howpublished={\url{https://huggingface.co/spaces/treble-technologies/ffasr}},
    note = {Accessed 2 September 2026}
}

@InProceedings{Panayotov2015,
  author =       {V. Panayotov and G. Chen and D. Povey and
                  S. Khudanpur},
  booktitle =    {Proc. ICASSP},
  title =        {Librispeech: An {ASR} corpus based on public domain
                  audio books},
  year =         2015
}

@InProceedings{Park2019,
  author =       {D. S. Park and W. Chan and Y. Zhang and C.-C. Chiu
                  and B. Zoph and E. D. Cubuk and Q. V. Le},
  title =        {{SpecAugment}: A Simple Data Augmentation Method for
                  Automatic Speech Recognition},
  booktitle =    {Proc. INTERSPEECH},
  year =         2019
}

@InProceedings{Pratap2020,
  author =       {V. Pratap and Q. Xu and A. Sriram and G. Synnaeve
                  and R. Collobert},
  title =        {{MLS}: A Large-Scale Multilingual Dataset for Speech
                  Research},
  booktitle =    {Proc. INTERSPEECH},
  year =         2020
}

@InProceedings{Saon2025,
  author =       {G. Saon and A. Dekel and A. Brooks and T. Nagano and
                  A. Daniels and A. Satt and A. Mittal and
                  B. Kingsbury and D. Haws and E. Morais and G. Kurata
                  and H. Aronowitz and I. Ibrahim and J. Kuo and
                  K. Soule and L. Lastras and M. Suzuki and R. Hoory
                  and S. Thomas and S. Novitasari and T. Fukuda and
                  V. Sunder and X. Cui and Z. Kons},
  title =        {Granite-speech: open-source speech-aware {LLMs} with
                  strong {English} {ASR} capabilities},
  booktitle =    {Proc. ASRU},
  year =         2025
}

@Article{Srivastava2014,
  author =       {N. Srivastava and G. Hinton and A. Krizhevsky and
                  I. Sutskever and R. Salakhutdinov},
  title =        {Dropout: A Simple Way to Prevent Neural Networks
                  from Overfitting},
  journal =      {Journal of Machine Learning Research},
  year =         2014,
  volume =       15,
  number =       56,
  pages =        {1929--1958}
}

@InProceedings{Tang2024,
  author =       {C. Tang and W. Yu and G. Sun and X. Chen and T. Tan
                  and W. Li and L. Lu and Z. Ma and C. Zhang},
  title =        {{SALMONN}: Towards generic hearing abilities for
                  large language models},
  booktitle =    {Proc. ICLR},
  year =         2024
}

@InProceedings{Wang2021,
  title =        "{V}ox{P}opuli: A Large-Scale Multilingual Speech
                  Corpus for Representation Learning, Semi-Supervised
                  Learning and Interpretation",
  author =       "C. Wang and M. Rivi\`{e}re and A. Lee and A. Wu and
                  C. Talnikar and D. Haziza and M. Williamson and
                  J. Pino and E. Dupoux",
  booktitle =    {Proc. ACL-IJCNLP},
  year =         2021
}

@misc{abouelenin2025phi,
  title =        {Phi-4-Mini Technical Report: Compact yet Powerful
                  Multimodal Language Models via Mixture-of-{LoRAs}},
  author =       {A. Abouelenin and A. Ashfaq and A. Atkinson and
                  H. Awadalla and N. Bach and J. Bao and A. Benhaim
                  and M. Cai and V. Chaudhary and C. Chen and D. Chen
                  and D. Chen and J. Chen and W. Chen and Y.-C. Chen
                  and Y. Chen and Q. Dai and X. Dai and R. Fan and
                  M. Gao and M. Gao and A. Garg and A. Goswami and
                  J. Hao and A. Hendy and Y. Hu and X. Jin and
                  M. Khademi and D. Kim and Y. J. Kim and G. Lee and
                  J. Li and Y. Li and C. Liang and X. Lin and Z. Lin
                  and M. Liu and Y. Liu and G. Lopez and C. Luo and
                  P. Madan and V. Mazalov and A. Mitra and A. Mousavi
                  and A. Nguyen and J. Pan and D. Perez-Becker and
                  J. Platin and T. Portet and K. Qiu and B. Ren and
                  L. Ren and S. Roy and N. Shang and Y. Shen and
                  S. Singhal and S. Som and X. Song and T. Sych and
                  P. Vaddamanu and S. Wang and Y. Wang and Z. Wang and
                  H. Wu and H. Xu and W. Xu and Y. Yang and Z. Yang
                  and D. Yu and I. Zabir and J. Zhang and L. L. Zhang
                  and Y. Zhang and X. Zhou},
  howpublished = {\url{https://arxiv.org/abs/2503.01743}},
  year =         2025
}

@inproceedings{burchi2021efficient,
  title =        {Efficient conformer: Progressive downsampling and
                  grouped attention for automatic speech recognition},
  author =       {M. Burchi and V. Vielzeuf},
  booktitle =    {Proc. ASRU},
  year =         2021
}

@inproceedings{chen2024salm,
  title =        {{SALM}: Speech-augmented language model with
                  in-context learning for speech recognition and
                  translation},
  author =       {Z. Cheni and H. Huang and A. Andrusenko and
                  O. Hrinchuk and K. C. Puvvadahna and J. Li and
                  S. Ghoshr and J. Balam and B. Ginsburg},
  booktitle =    {Proc. ICASSP},
  year =         2024
}

@misc{grattafiori2024llama,
  title =        {The {Llama} 3 herd of models},
  author =       {A. Grattafiori and A. Dubey and A. Jauhri and
                  A. Pandey and A. Kadian and A. Al-Dahle and
                  A. Letman and A. Mathur and A. Schelten and
                  A. Vaughan and others},
  howpublished = {\url{https://arxiv.org/pdf/2407.21783}},
  year =         2024
}

@inproceedings{ko2017reverberation,
  author =       {T. Ko and V. Peddinti and D. Povey and M. L. Seltzer
                  and S. Khudanpur},
  title =        {A Study on Data Augmentation of Reverberant Speech
                  for Robust Speech Recognition},
  booktitle =    {Proc. ICASSP},
  year =         2017,
}

@misc{lucidrains,
    author = {P. Wang},
    title = {conformer},
    howpublished = {GitHub repository, \url{https://github.com/lucidrains/conformer}}
}

@inproceedings{ng2021pushing,
  title =        {Pushing the limits of non-autoregressive speech
                  recognition},
  author =       {E. Ng and C.-C. Chiu and Y. Zhang and W. Chan},
  booktitle =    {Proc. INTERSPEECH},
  year =         2021
}

@inproceedings{nozaki2021relaxing,
  title =        {Relaxing the Conditional Independence Assumption of
                  {CTC}-Based {ASR} by Conditioning on Intermediate
                  Predictions},
  author =       {J. Nozaki and T. Komatsu},
  booktitle =    {Proc. INTERSPEECH},
  year =         2021
}

@inproceedings{rekesh2023fast,
  title =        {Fast conformer with linearly scalable attention for
                  efficient speech recognition},
  author =       {D. Rekesh and N. R. Koluguri and S. Kriman and
                  S. Majumdar and V. Noroozi and H. Huang and
                  O. Hrinchuk and K. Puvvada and A. Kumar and J. Balam
                  and others},
  booktitle =    {Proc. ASRU},
  year =         2023
}

@misc{snyder2015musan,
  author =       {D. Snyder and G. Chen and D. Povey},
  title =        {{MUSAN}: A Music, Speech, and Noise Corpus},
  year =         2015,
  howpublished = {\url{https://arxiv.org/abs/1510.08484}}
}

@article{shi2026qwen3,
  title={Qwen3-asr technical report},
  author={Shi, Xian and Wang, Xiong and Guo, Zhifang and Wang, Yongqi and Zhang, Pei and Zhang, Xinyu and Guo, Zishan and Hao, Hongkun and Xi, Yu and Yang, Baosong and others},
  journal={arXiv preprint arXiv:2601.21337},
  year={2026}
}
\end{document}